\documentclass[letterpaper]{article} 
\usepackage[draft]{aaai25}  

\usepackage{amsmath,amsfonts,bm}

\def\eqref#1{equation~\ref{#1}}

\def\1{\bm{1}}

\DeclareMathAlphabet{\mathsfit}{\encodingdefault}{\sfdefault}{m}{sl}
\SetMathAlphabet{\mathsfit}{bold}{\encodingdefault}{\sfdefault}{bx}{n}

\def\gA{{\mathcal{A}}}

\def\gD{{\mathcal{D}}}

\def\gM{{\mathcal{M}}}

\def\gS{{\mathcal{S}}}

\newcommand{\E}{\mathbb{E}}

\usepackage{times}  
\usepackage{helvet}  
\usepackage{courier}  
\usepackage[hyphens]{url}  
\usepackage{graphicx} 
\usepackage{natbib}  
\usepackage{caption} 
\usepackage{newfloat}
\usepackage{listings}
\DeclareCaptionStyle{ruled}{labelfont=normalfont,labelsep=colon,strut=off} 

\usepackage{amsmath}
\usepackage{amssymb}
\usepackage{amsthm}
\usepackage{mathtools}
\usepackage{algorithm}
\usepackage{algpseudocode}
\usepackage{lipsum}

\usepackage[colorlinks=true,linkcolor=blue,
anchorcolor=blue,
citecolor=blue,backref=page]{hyperref}       
\usepackage{nameref}
\usepackage{cleveref}

\Crefname{equation}{Eqn.}{Eqns.}
\Crefname{figure}{Fig.}{Figs.}
\Crefname{tabular}{Tab.}{Tabs.}
\Crefname{section}{Section}{Sections}

\usepackage{url}            
\usepackage{booktabs}       
\usepackage{amsfonts}       
\usepackage{nicefrac}       
\usepackage{microtype}      
\usepackage{xcolor}         
\usepackage{colortbl}
\usepackage{subcaption}

\usepackage{array}
\usepackage{bm}

\usepackage{tabularx}
\usepackage{graphicx}
\usepackage{changepage}

\usepackage{float}

\theoremstyle{plain}

\theoremstyle{definition}

\theoremstyle{remark}

\title{Are Expressive Models Truly Necessary for Offline RL?}
\author{
    Guan Wang\equalcontrib\textsuperscript{\rm 1}, Haoyi Niu\equalcontrib\textsuperscript{\rm 1}, Jianxiong Li\textsuperscript{\rm 1}, Li Jiang\textsuperscript{\rm 2}, Jianming Hu\textsuperscript{\rm 1\dag}, Xianyuan Zhan\textsuperscript{\rm 1,3,4}\thanks{Corresponding authors. 
    Code is available at \href{https://github.com/imoneoi/RSP_JAX}{RSP\_ JAX}.
    }
}
\affiliations {
    \textsuperscript{\rm 1}Tsinghua University, \textsuperscript{\rm 2}McGill University, \textsuperscript{\rm 3}Shanghai AI Laboratory, \textsuperscript{\rm 4}Beijing Academy of Artificial Intelligence\\
    \texttt{imonenext@gmail.com,\{nhy22@mails,hujm@mail,zhanxianyuan@air\}.tsinghua.edu.cn}
}

\begin{document}

\maketitle

\begin{abstract}
Among various branches of offline reinforcement learning (RL) methods, goal-conditioned supervised learning (GCSL) has gained increasing popularity as it formulates the offline RL problem as a sequential modeling task, therefore bypassing the notoriously difficult credit assignment challenge of value learning in conventional RL paradigm. Sequential modeling, however, requires capturing accurate dynamics across long horizons in trajectory data to ensure reasonable policy performance. To meet this requirement, leveraging large, expressive models has become a popular choice in recent literature, which, however, comes at the cost of significantly increased computation and inference latency. Contradictory yet promising, we reveal that lightweight models as simple as shallow 2-layer MLPs, can also enjoy accurate dynamics consistency and significantly reduced sequential modeling errors against large expressive models by adopting a simple recursive planning scheme: recursively planning coarse-grained future sub-goals based on current and target information, and then executes the action with a goal-conditioned policy learned from data relabeled with these sub-goal ground truths. We term our method as \textbf{R}ecursive \textbf{S}kip-Step \textbf{P}lanning (\textbf{RSP}). Simple yet effective, RSP enjoys great efficiency improvements thanks to its lightweight structure, and substantially outperforms existing methods, reaching new SOTA performances on the D4RL benchmark, especially in multi-stage long-horizon tasks. 

\end{abstract}

\section{Introduction}

\begin{figure}[t!]
    \centering
    \includegraphics[width=\linewidth]{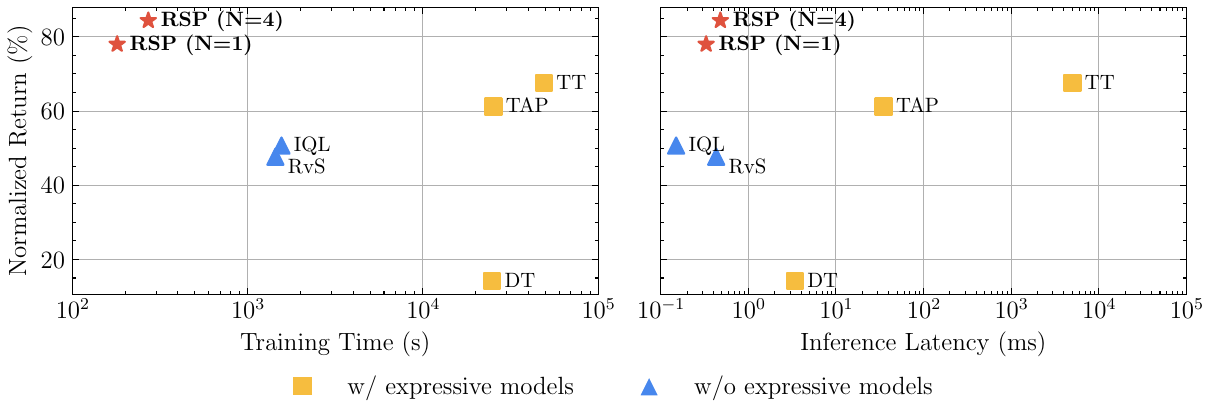}
    \caption{\small\textbf{Existing GCSL methods trade off score performances for slow training and inference.} However, RSP with recursion depth $N=1$ significantly outperforms all baselines on Antmaze datasets while completing training in just 180 seconds and enjoying an inference latency of 0.33ms. $N=4$ further enhances performance with minimal cost, as the inference time increases to 0.48 ms and the training time to 271s, remaining orders of magnitude lower than other methods that use expressive models.}
    \label{fig:plot_performance}
\end{figure}

Offline reinforcement learning (RL) has emerged as a promising approach to solving many real-world tasks using logged experiences without extra costly or unsafe interactions with environments~\citep{fujimoto2019off,levine2020offline, zhan2022deepthermal}. 
Unfortunately, the absence of online interaction also poses the counterfactual reasoning challenge in out-of-distribution (OOD) regions, causing catastrophic failures in the conventional RL paradigm due to extrapolation error accumulation during value function learning using temporal difference (TD) updates~\citep{fujimoto2019off, kumar2019stabilizing,li2022data}.


To remedy this, Goal-Conditioned Supervised Learning (GCSL) has gained much attention as an alternative paradigm since it reformulates offline RL problems as sequential modeling tasks, where policy extraction can be learned in a supervised manner, thus bypassing problematic value update in conventional TD learning.
However, this comes with the price of much higher requirements of accurately describing the data dynamics in sequential modeling, which naturally favors more expressive models.
Recent advances in highly expressive models have yielded remarkable achievements across a diverse range of domains such as computer vision~\citep{liu2021swin} and natural language processing~\citep{vaswani2017attention}, which
have also been extended to offline RL to accurately capture policy distributions~\citep{wang2022diffusion, hu2023instructed, hansen2023idql, ada2023diffusion, wang2023offline, chen2022offline}, or reduce the accumulative compounding errors in sequential data modeling~\citep{chen2021decision, janner2021offline, janner2022planning, ajay2022conditional, villaflor2022addressing, jiang2022efficient, mazoure2023value,niu2024xted}. 
Although they sometimes provide encouraging improvements over previous approaches, they also suffer from noticeable performance deterioration when they fail to accurately represent the behavior policy and/or environment dynamics in long-horizon tasks. 

Moreover, the integration of expressive models in offline RL inevitably increases both the computational load and inference latency. This poses a significant challenge for many real-world applications that are latency-sensitive or resource-constrained. Besides, methods that employ such expressive models are notoriously data-hungry to train, making them impractical for scenarios with expensive samples \citep{zhan2022deepthermal,cheng2024look}. In Fig.~\ref{fig:plot_performance}, we compare key metrics of recent offline RL methods, including policy performance, training time, and inference latency. The use of large expressive models in prior offline RL methods such as DT~\citep{chen2021decision}, TAP~\citep{zhang2022efficient}, TT~\citep{janner2021offline}, and Diffuser~\citep{janner2022planning} sometimes results in marginal performance gains. However, this incremental improvement is offset by the exponentially increased computational load and inference latency, particularly when compared to models that do not employ such expressive models. This prompts us to ask the question: 
\begin{center}
\emph{Are expressive models truly necessary for offline RL?}
\end{center}


While prior attempts introduce highly expressive models to combat the accumulated compounding error in long-sequence modeling, these methods still operate in a \textit{fine-grained} (step-by-step) manner~\citep{chen2021decision, wang2022diffusion, janner2022planning}, which is inherently tied to the temporal structure of the environment and is susceptible to rapidly accumulated approximation errors over longer horizons~\citep{levine2020offline}. In this study, we provide a novel perspective to avoid step-wise accumulated compounding error in sequential modeling while only relying on a lightweight model architecture (as simple as 2-layer MLPs). The core of our method is the Recursive Skip-step Planning (RSP) scheme, which employs an iterative two-phase approach to solve tasks: it recursively plans skip-step future sub-goals conditioned on the current and target information, and then executes a goal-conditioned policy based on these coarse-grained predictions. During the recursive planning phase, RSP bypasses step-wise sub-goal prediction and focuses more on longer-horizon outcomes through recursively skipping steps. In essence, RSP can generate long-horizon sub-goals with just a few planning steps, which uses exponentially fewer steps than what is required by previous fine-grained sequence modeling methods, therefore smartly bypassing the long-horizon compounding error issue while enjoying great computational efficiency.

RSP can be easily implemented using simple two-layer shallow MLPs for recursive skip-step dynamics models and the goal-conditioned policy. The entire learning process can be conducted in a supervised learning fashion, eliminating the need for complex stabilization tricks or delicate hyperparameter tuning in value learning. Notably, RSP not only exhibits great training efficiency but also enjoys very low inference complexity, while it achieves comparable or even superior performance against existing offline RL algorithms on D4RL benchmark~\citep{fu2020d4rl}, including those equipped with expressive models. This advantage is particularly evident in complex tasks such as \texttt{AntMaze} and \texttt{Kitchen}, demonstrating the effectiveness of the long-horizon modeling capability by adopting our recursive skip-step planning scheme. 

\section{Related Work}
\textbf{Model-free Offline RL.}
Most prior methods incorporate pessimism during training to alleviate the distributional shift problem~\citep{levine2020offline}. One solution is leveraging various behavior regularizations to constrain the learned policies close to the behavior policy in offline dataset~\citep{fujimoto2019off, wu2019behavior, kumar2019stabilizing, fujimoto2021minimalist, li2022data}. Some other works introduce pessimism during policy evaluation by assigning low confidences or low values for the value functions in out-of-distribution (OOD) regions~\citep{kumar2020conservative, kostrikov2021offline, lyu2022mildly, an2021uncertainty,bai2021pessimistic,niu2022trust}. In-sample learning methods~\citep{kostrikov2021iql, xu2023offline, xu2022policy, garg2023extreme, xiao2023sample, wang2024offline, zhengsafe} have recently emerged as an alternative, optimizing on samples exclusively from the offline dataset and thus eliminating any OOD value queries. Moreover, an alternative in-sample learning approach performs RL tasks via supervised learning by conditioning other available sources, also called Reinforcement Learning via Supervised Learning) (RvS) \citep{kumar2019reward, schmidhuber2019reinforcement, emmons2021rvs, feng2022curriculum}. These approaches are super stable and easy to tune compared to other methods. 

Existing methods, however, rely on shallow MLPs to model their actors, performing well on simple tasks but falling short on more complex long-horizon planning tasks, such as \texttt{AntMaze} and \texttt{Kitchen} tasks. To combat this, one popular direction is to increase policy capacity by employing highly expressive models to accurately approximate the actor, which obtains certain degrees of performance gains~\citep{wang2022diffusion, hansen2023idql, ada2023diffusion, chen2022offline, lu2023contrastive}. However, this marginal improvement comes at the cost of extra model complexity and exponentially increased computational burden, inevitably limiting their applications.

\textbf{Sequential Modeling for Offline RL.} 
Different from traditional TD methods, this avenue treats offline RL problems as general sequence modeling tasks, with the motivation to generate a sequence of actions or trajectories that attain high rewards. In this formulation, shallow feedforward models are typically believed to suffer from a limited modeling horizon due to accumulated modeling errors caused by deficient model capacities~\citep{janner2019trust, amos2021model,zhan2022model}. To combat this, recent innovations in sequential modeling techniques shift towards achieving precise long-horizon modeling through the use of high-capacity sequence model architectures such as Transformer \citep{chen2021decision, janner2021offline, jiang2022efficient, wang2022bootstrapped, konan2023contrastive, hu2023graph, wu2023elastic, villaflor2022addressing} and diffusion models~\citep{janner2022planning, ajay2022conditional,niu2024xted}. 
However, Fig.~\ref{fig:plot_performance} manifests that the adoption of expressive models imposes a large amount of computational load and complexity. 
In this study, RSP adopts a recursive coarse-grained paradigm for sub-goal prediction solely with shallow 2-layer MLPs, achieving exceptional results while bypassing training inefficiency and high inference latency issues led by expressive models.

\section{Preliminaries}
\label{sec: prelims}

In this paper, we consider the standard Markov Decision Process (MDP), denoted by $\gM=(\gS, \gA, T, \rho, r, \gamma)$, which is characterized by states $s \in \gS$, actions $a \in \gA$, transition probabilities $T(s_{t+1}|a_t, s_t)$, initial state distribution $\rho(s_0)$, reward function $r$ and the discount factor $\gamma \in (0, 1)$. In each episode, the agent is given a reward function $r(s_t, a_t)$ and embodies with a policy $\pi(a \mid s)$ to interact with the environment. Let $\tau := (s_0, a_0, s_1, a_1, \cdots)$ denote an infinite length trajectory. We employ $\pi(\tau)$ to represent the probability of sampling trajectory $\tau$ from the policy $\pi(a \mid s)$. The primary goal is to optimize the expected cumulative discounted rewards incurred along a trajectory, where the objective and the corresponding Q-functions are expressed as:
\begin{equation}
\begin{aligned}
    &\max_\pi \; \mathbb{E}_{\pi(\tau)} \Big[\sum_{t=0}^\infty \gamma^t r(s_t, a_t) \Big],\\ 
    &Q^{\pi}(s, a) := \mathbb{E}_{\pi(\tau)}  \bigg[\sum_{t=0}^\infty \gamma^t r(s_t, a_t) \bigg \vert \substack{s_0 = s\\ a_0 = a}\bigg].
\end{aligned}
\label{eq:Q}
\end{equation}
We are interested in the offline setting in this work and the objective of offline RL is in line with traditional RL, but with infeasible additional online interactions. In the offline setting, a fixed dataset is given by $\gD := \{\tau_1, \tau_2, \tau_3, \cdots\}$, which comprises multiple trajectories gathered from unknown behavior policies, denoted with $\beta(a|s)$ in this work. 


\subsection{RL via Goal-Conditioned Supervised Learning}

\begin{algorithm}[H]
    \caption{\footnotesize General Procedure of GCSL}\label{alg:rvs}
    \begin{algorithmic}[1] \footnotesize 
        \State \textbf{Input}: Offline dataset $\gD$
        \State $\textsc{Relabel}(\tau)$, $\forall\tau \sim \gD$ \Comment{Step 1}
        \State Sample transitions $(s, a, e) \in \gD$
        \State Update $\pi(a \mid s, e)$ with \Cref{eq:rvs-obj}  \Comment{Step 2} 
        \State \textbf{return} $\pi(a \mid s, e)$
    \end{algorithmic}
\end{algorithm}

Reformulating RL as GCSL, previous works~\citep{schmidhuber2019reinforcement, emmons2021rvs} typically employ hindsight relabeling to perform conditional imitation learning. 
The core idea underlying this approach is that any trajectory can be considered a successful demonstration of reaching certain goals, i.e. future states or accumulating rewards within the same trajectory. 
GCSL demonstrates significant potential in the offline setting, largely due to the availability of that goal information \citep{kumar2019reward, chen2021decision, emmons2021rvs, feng2022curriculum}. 

The general procedure of GCSL is simple and straightforward, as shown in~\ref{alg:rvs}, consisting of two main steps. 
In the first step, GCSL relabels the dataset by adding an additional goal information $e$ to each state-action pair and forms a state-action-goal tuple, i.e., $(s, a, e)$. 
The goal can be the target state, cumulative return, or language description \citep{lynch2020learning, ghosh2020learning, kumar2019reward, chen2021decision, feng2022curriculum}, if applicable. 
For the state $s_t$, the goal $e_t$ is randomly sampled from the future steps from the current step $t$ along a valid trajectory in the dataset. The key idea behind this resampling is to consider any future goal $e_t$ is reachable by taking action $a_t$ and state $s_t$. In the second step, GCSL learns a goal-conditioned policy within the relabeled dataset via maximum log-likelihood: 
\begin{align}
    \arg \max_{\pi}\E_{(s, a, e) \sim \gD}\left[\log \pi_{\phi}(a \mid s, e) \right]. \label{eq:rvs-obj}
\end{align}

\section{Recursive Skip-Step Planning}


\subsection{Why Non-recursive Planning Might Struggle?}
Fine-grained planning using single-step dynamics \( f(s_{t+1}|s_t,g) \) often faces significant challenges due to the rapid accumulation of step-wise errors in sequential modeling, led by such a limited planning horizon.
Although expressive models like Transformers~\citep{chen2021decision,janner2021offline} and Diffusers~\citep{janner2022planning,ajay2022conditional} aim to mitigate these issues by modeling long-term dependencies, their complex architectures significantly increase computational load. This approach ultimately struggles to balance prediction accuracy with efficiency, remaining constrained within the limitations of fine-grained sequential modeling.

To overcome this, coarse-grained planning uses skip-step dynamics, predicting intermediate sub-goals via \( f(s_{t+k} | s_t, g) \) with an extended planning horizon $k$. However, setting an appropriate skip-step horizon \( k \) is non-trivial: too short a horizon fails to alleviate error accumulation as in fine-grained planning, while too long a horizon weakens the learning signal, compromising policy performance in long sequential tasks. Thus, a delicate balance must be achieved between the prediction stability of the sub-goal \( s_{t+k} \) and the control precision of the policy \( \pi(a_t | s_t, s_{t+k}, g) \). Non-recursive planning may struggle to select sub-goal information that provides both adequate long-term guidance from the final goal and sufficient supervision for current action.

\subsection{Coarse-grained Planning in a Recursive Way}

\begin{figure}[t!]
    \vspace{-0.3cm}
    \centering
    \includegraphics[width=.8\linewidth]{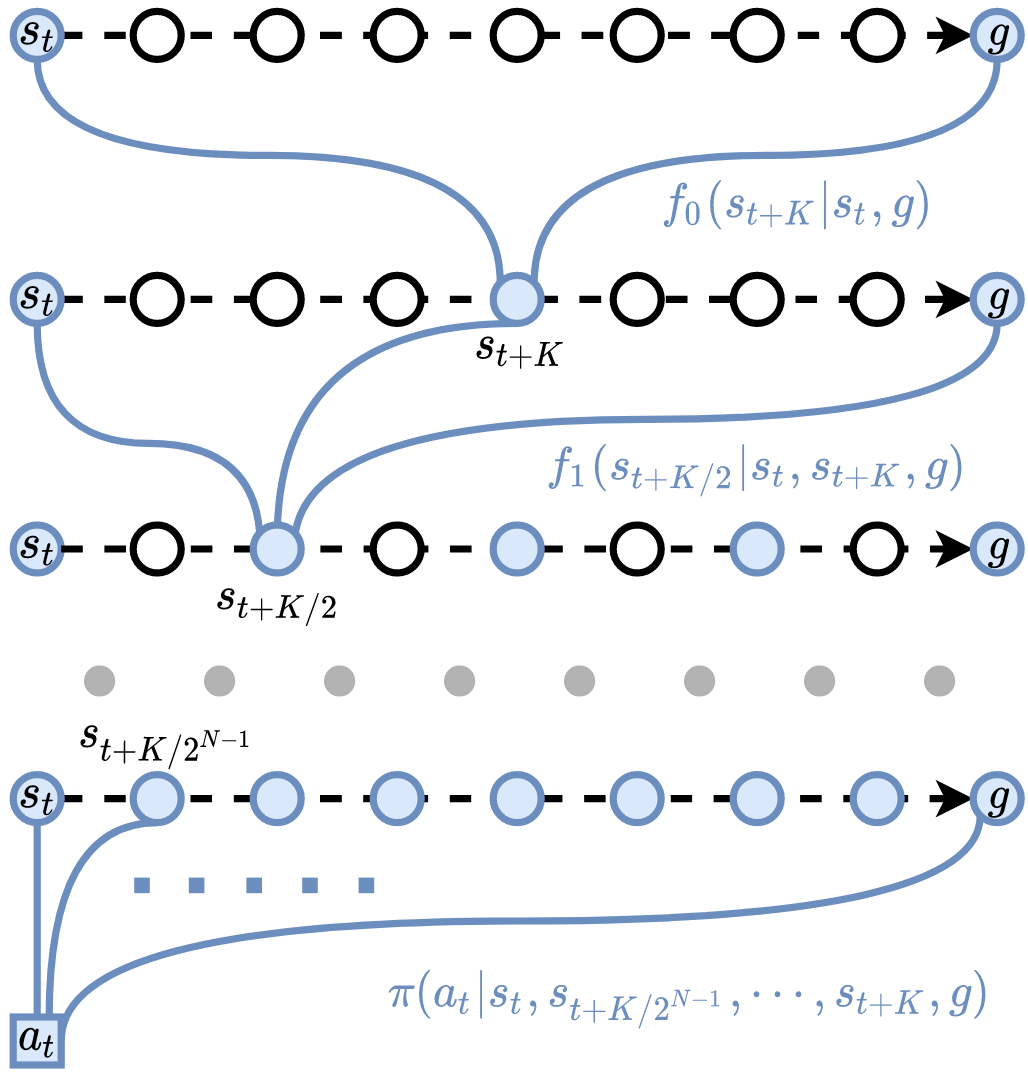}
    \caption{The illustration of Recursive Skip-Step Planning}
    \label{fig:rsp}
    \vspace{-0.3cm}
\end{figure}

The dilemma for non-recursive planning suggests using a higher-level dynamics model to predict distant sub-goals \( f'(s_{t+k'}|s_t, g) \) and conditioning intermediate predictions on the current-level dynamics \( f(s_{t+k}|s_t, s_{t+k'}, g) \).
This hierarchical approach stabilizes sub-goal predictions $\hat{s}_{t+k}$ in the same way that sub-goals enhance action extraction $\hat{a}_t$. 
By recursively introducing next-level sub-goal prediction with more distant horizons, sub-goal approximation error is reduced at each level.
As these sub-goals approach the final target, their error also inherently diminishes.
Planning in such a recursive manner thus effectively mitigates error accumulation in long-horizon tasks.

In the RSP framework, each current state \( s_t \) in the dataset is first relabeled with a fixed-horizon future target state \( s_{t+k} \) as the sub-goal, resulting in an augmented dataset $\mathcal{D}=\{(s_t, a_t, s_{t+k},g)\}$. This allows us to revise the optimization objective of goal-conditioned policies in Equation~\ref{eq:rvs-obj} to focus on goal-conditioned sub-goal prediction:
\begin{equation}
    \hat{f}=\underset{f}{\arg\max}\ \E_{(s_t, s_{t+k}, g) \sim \gD}\left[\log f(s_{t+k}|s_t, g) \right]. \label{eq:1_plan}
\end{equation}
As explained in the previous section, planning with a learned coarse-grained dynamics model \( \hat{f} \) is insufficient for balancing the minimization of action and sub-goal prediction errors. To further reduce the sub-goal prediction error, we extend it to a recursive skip-step prediction strategy, as depicted in Fig.~\ref{fig:rsp}, predicting high-level sub-goals to stabilize the lower-level sub-goal predictions in sequential modeling.
To standardize the recursive process, we set $K^{(n)}:=t+K/2^{n}, n\in[0,\cdots,N-1]$ where $N$ is the recursion depth, $K$ is the skip step of the highest-level sub-goal, and $s_{K^{(n)}}$ are sub-goal ground truths. 
The highest-level sub-goal skips $K$ steps while the lowest-level sub-goal skips $k=K/2^{N-1}$ steps from the current step.
For $N$-level recursion, we need to relabel the dataset $\gD=\{(s_t,a_t,g)\}$ with additional $N$ skip-step sub-goal ground truths:
\begin{equation}
    \gD\leftarrow\left\{\left(a_t,\kappa^{(N)}_t:=(s_t, s_{K^{(N-1)}},\cdots, s_{K^{(0)}}, g)\right)\right\}\label{eq:relabel}
\end{equation}
where \(\kappa_t^{(0)}=(s_t,g);\ \forall n \in [1, \dots, N]\), \(\kappa^{(n)}_t=s_{K^{(n-1)}}\cup\kappa^{(n-1)}_t\) contains all the information required by the \(n\)-th level dynamics model to fit the sub-goal ground truth \(s_{K^{(n-1)}}\). 
For brevity, we omit \(t\) and refer to \(\kappa^{(n)}\) in the following descriptions.
With all the definitions above, the highest(first)-level coarse-grained dynamics model could be denoted by $\hat{f}_0(s_{K^{(0)}}|s_t, g)$.
Subsequently, we can learn more coarse-grained dynamics models to push the limits of the benefits of coarse-grained planning, as depicted in Fig.~\ref{fig:rsp}. In terms of the $n$-th dynamics model, the recursion solution naturally extends to the optimization objective in Eq.~\ref{eq:n_plan}:
\begin{equation}
\small
    \underset{f_{n}}{\max}\ \E_{\kappa^{(n)}\sim\gD}\left[\log f_{n}(s_{K^{(n-1)}}|\kappa^{(n-1)}) \right], \forall n \in [1, \dots, N]. \label{eq:n_plan}
\end{equation}
where the current level sub-goal generation is conditioned on all the sub-goals predicted with higher-level dynamics for stability, which proves a significant design in our experiments. 

During the evaluation process, we can formulate the recursive sub-goal planning as:
\begin{equation}
\small
    \widehat{s_{K^{(n-1)}}} = \underset{s_{K^{(n-1)}}}{\arg\max}\ \hat{f}_{n}(s_{K^{(n-1)}}| \widehat{\kappa^{(n-1)}}), \forall n\in[1,\cdots, N]
    \label{eq: eval_dynamics_iter}
\end{equation}
where \(\widehat{\kappa^{(0)}}=\kappa^{(0)}=(s_t,g);\ \forall n \in [1, \dots, N]\), \(\widehat{\kappa^{(n)}}=\widehat{s_{K^{(n-1)}}}\cup\widehat{\kappa^{(n-1)}}\).
Eventually, we get the predicted sub-goals and leverage $\widehat{\kappa^{(N)}}$ for policy extraction.

\begin{algorithm}[t]
    \caption{\footnotesize Recursive Skip-Step Planning (RSP) }\label{alg:rsp}
        \begin{algorithmic}[1] \footnotesize
        \State \textbf{Input}: Dataset $\gD$, Recursion depth $N$, Fixed planning step $K$
        \State Relabel Dataset with Eq.~\ref{eq:relabel}\Comment{Step 1}
        \State \texttt{\color{blue}{\textbf{// Training}}}\Comment{Step 2}
        \For{every training step}
            \State Sample transitions $(s_t, a_t, s_{K^{(0)}},\cdots, s_{K^{(N-1)}}, g) \in \gD$
            \For{$n=1,\cdots,N$}
                \State Update the $n$-th dynamics model $\hat{f}_n$ by Eq.~\ref{eq:n_plan} 
            \EndFor
            \State Update $\pi_\phi$ by Eq.~\ref{eq:rsp_policy}  
        \EndFor
        \State \texttt{\color{blue}{\textbf{// Evaluation}}}\Comment{Step 3}
                \State Get initial state $s$, set \texttt{done} as False
                \While {not \texttt{done}}
                \State Get action $a$ from Eq.~\ref{eq: eval}
                \State Roll out $a$ and get $(s^{\prime}, r, \texttt{done})$
                \State Set $s=s'$
                \EndWhile
        \end{algorithmic}
\end{algorithm}

\subsection{Policy Extraction}
\label{subsec: policy_rsp}

    

Different from other planning methods, the coarse-grained planning approach employed by RSP does not explicitly integrate the corresponding policy into the learning process, necessitating a separate policy extraction step. To maintain the simplicity and efficiency of RSP, we extract the policy by maximizing the log-likelihood within a supervised learning framework. We introduce a goal-conditioned policy, distinct in that it incorporates all the sub-goals planned from prior coarse-grained dynamics models:
\begin{align}
     \underset{\phi}{\max}\ \E_{(a_t, \kappa^{(N)})\sim\gD}\left[\log \pi_{\phi}
    (a_t \mid  \kappa^{(N)}) \right]. \label{eq:rsp_policy} 
\end{align}
At the evaluation stage, given the current state $s_t$, the action is determined by both the coarse-grained dynamics model and the goal-conditioned policy with:
\begin{equation}
    \widehat{a_t} = \underset{a_t}{\arg\max}\ \hat{\pi}_\phi(a_t|\widehat{\kappa^{(N)}}))
    \label{eq: eval}
\end{equation}
where $\widehat{\kappa^{(N)}}=(s_t, \widehat{s_{K^{(N-1)}}},\cdots, \widehat{s_{K^{(0)}}}, g)$ is obtained from Eq.~\ref{eq: eval_dynamics_iter} recursively.


The final procedure in Algorithm~\ref{alg:rsp} consists of three steps. Initially, it relabels the dataset $\mathcal{D}$, augmenting each transition with the future skip-step sub-goal and final goal ground truths. Subsequently, RSP learns $N$ coarse-grained dynamics models from Eq.~\ref{eq:n_plan} and a goal-conditioned policy from Eq.~\ref{eq:rsp_policy} that determines the action based on the sub-goals predicted with all the dynamics models. Finally, RSP plans the sub-goal sequence recursively with Eq.~\ref{eq: eval_dynamics_iter} and executes action from sub-goals with Eq.~\ref{eq: eval}.

\begin{figure*}[t]
    \centering
        \begin{subfigure}[t]{0.19\textwidth}
            \centering
            \includegraphics[width=\textwidth]{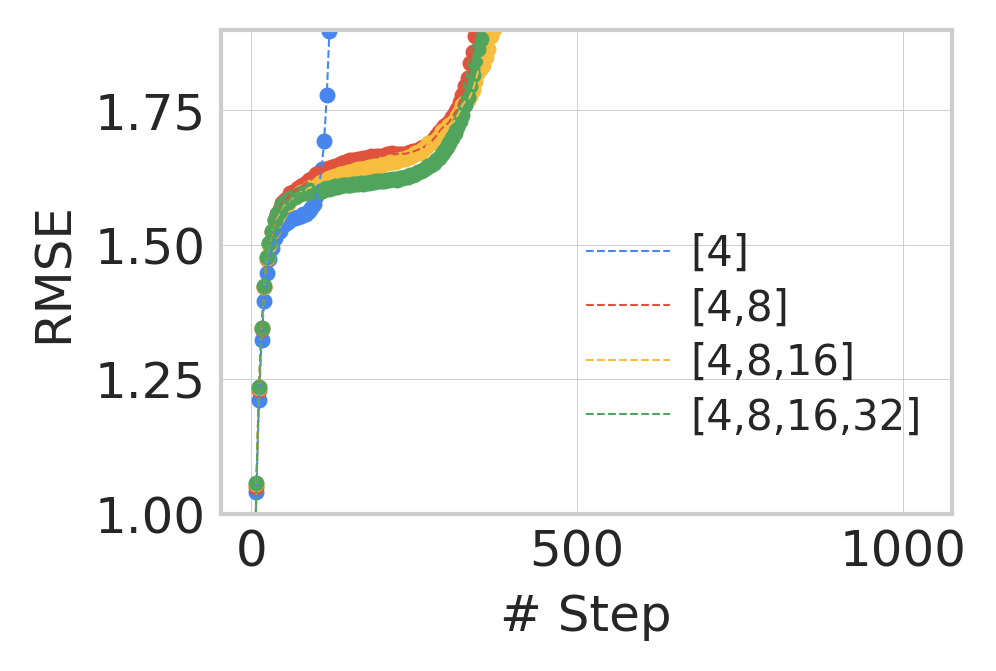}
            \label{fig:pred_erros_4_main}
        \end{subfigure}
        \hfill
        \begin{subfigure}[t]{0.19\textwidth}
            \centering
            \includegraphics[width=\textwidth]{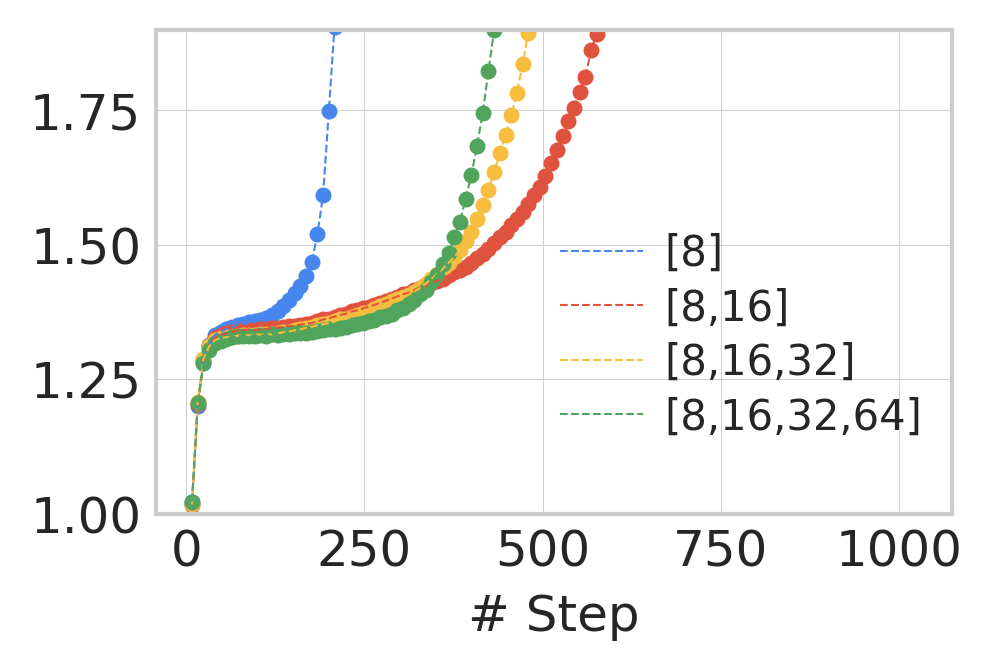}
            \label{fig:pred_error_8_main}
        \end{subfigure}
        \hfill  
        \begin{subfigure}[t]{0.19\textwidth}
            \centering
            \includegraphics[width=\textwidth]{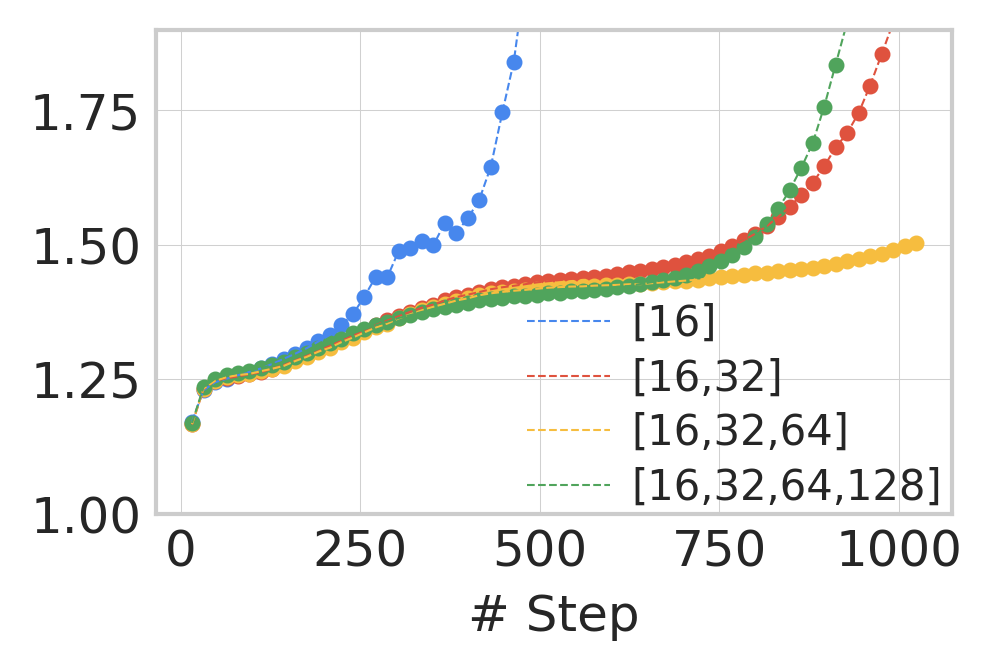}
            \label{fig:pred_error_16_main}
        \end{subfigure}
        \hfill  
        \begin{subfigure}[t]{0.19\textwidth}
            \centering
            \includegraphics[width=\textwidth]{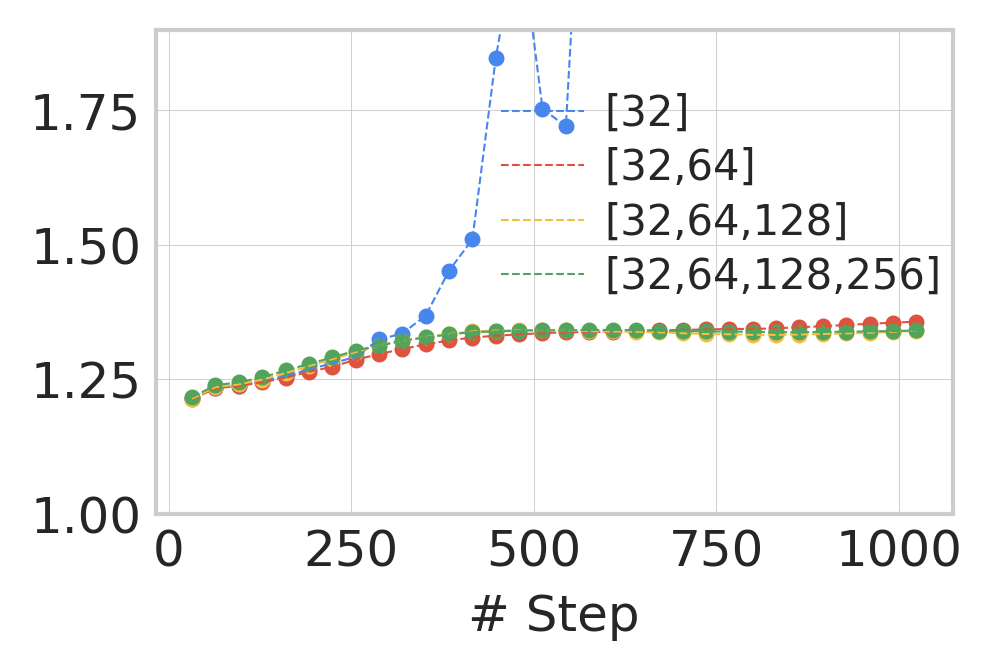}
            \label{fig:pred_error_32_main}
        \end{subfigure}
        \hfill  
        \begin{subfigure}[t]{0.19\textwidth}
            \centering
            \includegraphics[width=\textwidth]{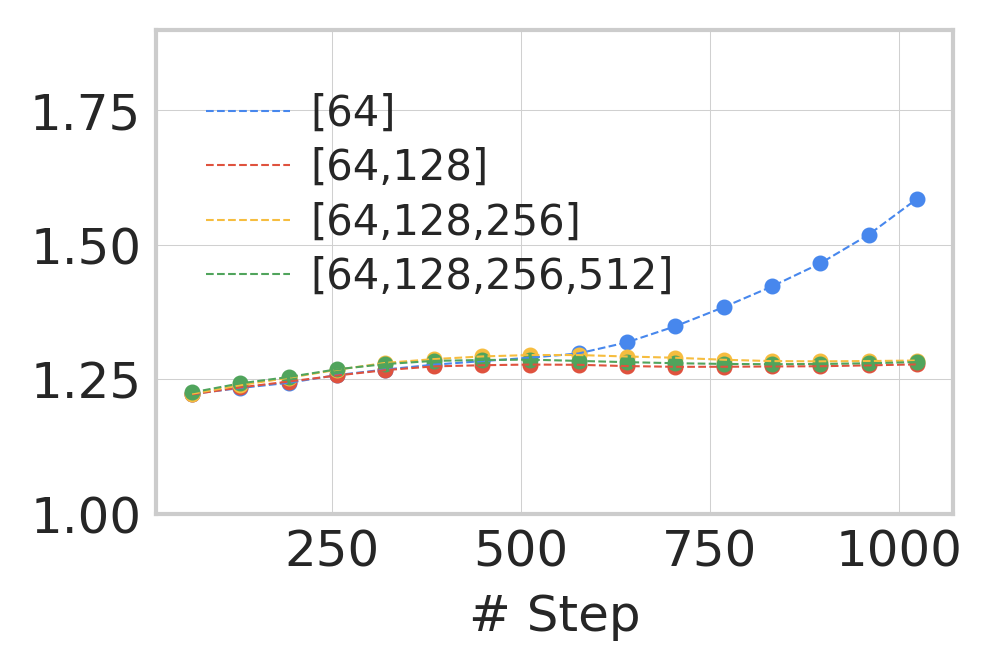}
            \label{fig:pred_error_64_main}
        \end{subfigure}
        \vspace{-5mm}
        \caption{Root mean squared error (RMSE) between rollout sub-goal predictions and ground truths on \texttt{Antmaze-Ultra}.
        }
    \label{fig:pred_error_main}
\end{figure*}

\subsection{Discussions}\label{sec:discuss}
 In this section, we provide an in-depth explanation of why recursive skip-step planning might outperform from the perspective of model rollout prediction errors. We conduct experiments on different (lowest-level) skip-step horizons ($k=4,8,16,32,64$) and recursion depths ($N=1,2,3,4$) in the long-horizon \texttt{Antmaze-Ultra-Diverse} task, where trajectories can extend up to 3000 steps. 
For example, [16,32,64] in Fig.~\ref{fig:pred_error_main} denotes $N=3$ and $k=16$.
We plot the root mean squared errors (RMSE) between the model rollout predictions and the ground truths of skip-step state trajectories, averaged over 10 seeds. 
Specifically, we obtain the next skip-step prediction $\hat{s}_{t+k}$ through the recursive approach of RSP; we iteratively query the next skip-step prediction $\hat{s}_{t+nk}$ from the last prediction $\hat{s}_{t+(n-1)k}$ and finally obtain rollout sub-goal predictions $s_t, \cdots, \hat{s}_{t+(n-1)k}, \hat{s}_{t+nk}, \cdots, \hat{s}_{t+1024}$; then calculate RMSE between the predictions and ground truths. 

In Fig.~\ref{fig:pred_error_main}, as $k$ increases from 4 to 64, the accuracy of the model rollout prediction improves, aligning more closely with the ground truth over longer-horizon rollouts. This suggests that coarse-grained prediction outperforms fine-grained prediction in mitigating accumulated compounding errors as the skip-step horizon $k$ increases, thereby enhancing long-horizon planning capabilities. 
Besides, as the recursion depth $N$ increases, we observe that the RMSE of model rollout predictions achieves a lower asymptotic value and remains closer to the ground truth over longer-horizon rollouts, demonstrating the effectiveness of recursive planning.

\noindent\textbf{Choices of recursion depth and skip-step horizon.}\ At first glance, Fig.~\ref{fig:pred_error_main} indicates that increasing recursion depth and skip-step horizon improves the planning performance of RSP. However, these plots assume that an oracle policy model perfectly follow the skip-step state predictions. In practice, deviations from optimal state trajectories can easily occur due to suboptimal policies, especially when the policy model $\pi(a_t|s_t,s_{t+k},\cdots,g)$ is conditioned on a distant skip-step state prediction $s_{t+k}$, which offers minimal guidance for current actions.
Therefore, selecting recursion depth and horizon requires to strike the delicate balance between the expressiveness of policy models and the long-horizon capability of dynamics models. Given that we use a 2-layer MLP for policy learning for simplicity in experiments, we opt for $k=32$ and $N=1$ in recursive skip-step dynamics models across all tasks in order to provide sufficient long-horizon planning ability while easing the burden for policy learning. Practitioners using more advanced policy modeling and learning methods may consider enlarging the horizon and deepening the recursion process accordingly.

\begin{figure*}[t]
    \vspace{-0.3cm}
    \centering
    \includegraphics[width=1.0\linewidth]{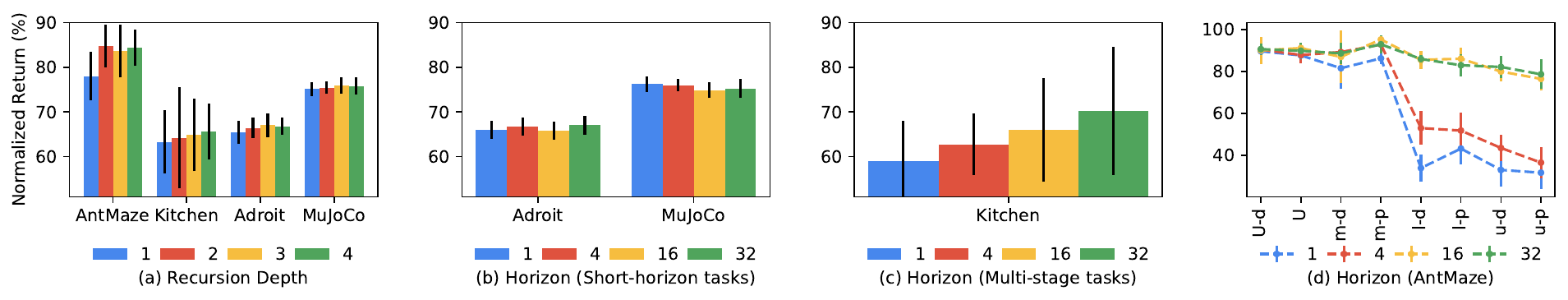}
    \caption{Ablations on choices of recursion depth and horizon. }
    \label{fig:ablation}
\end{figure*}

\section{Experiments}
\label{sec:exp}
In this section, we conduct comprehensive evaluations to showcase the effectiveness of RSP. Specifically, 
we compare RSP against other competitive baselines on the standard offline RL benchmark, D4RL~\citep{fu2020d4rl}. The results indicate that despite its simplicity,  RSP can obtain on-par or superior performance compared to others using expressive models, especially on the long-horizon \texttt{Antmaze} and multi-stage \texttt{Kitchen} tasks (Section \textit{\nameref{sec:d4rl_benchmarks}}). Moreover, RSP exhibits fast training and inference time, with training for only around 180 seconds and an inference latency of under 1ms (Section \textit{\nameref{sec:cost_latency}}). At last, we provide extensive ablation studies on some critical components of RSP including the planning horizon and recursion depth (Section \textit{\nameref{sec:ablation}}). 



\subsection{D4RL Benchmark Results}\label{sec:d4rl_benchmarks}
\subsubsection{Evaluation setups}
We evaluate the performance of various methods using a diverse set of challenging tasks from D4RL benchmark~\citep{fu2020d4rl}, including \texttt{Antmaze navigation}, \texttt{Kitchen Manipulation}, \texttt{Adroit hand manipulation}, and  \texttt{Mujoco locomotion} tasks. The Antmaze navigation tasks require intricate long-horizon planning to guide an ant navigate from the starting point to the goal location, which are pretty challenging for traditional offline RL to solve. To test the planning capabilities on extremely long-horizon tasks ($>2000$ steps), we consider a more extensive and challenging \texttt{Ultra} Antmaze environment. 
Kitchen tasks involve a 9-DoF Franka robot interacting with a kitchen scene to complete a combination of household sub-tasks, which requires multi-skill long-horizon composition ability.
In contrast, the Adroit hand manipulation tasks involve controlling a dexterous robot hand to complete tasks like opening a door or rotating a pen, with planning horizons typically ranging from 100 to 200 steps, which are relatively simple. Finally, the Mujoco locomotion tasks are believed as the simplest tasks among all these tasks. They only focus on controlling different robots to move forward as fast as possible, which are relatively simple since the control pattern is repetitive. To assess performance, we normalize the final scores by following the benchmark standards~\citep{fu2020d4rl}, where 0 represents random policy performance and 100 means expert-level performance. We use 10 random seeds and 100 episodes per seed for every task in our evaluation, with mean and standard deviation reported. 
All the results are obtained on a server with 512G RAM, $4\times$ A800 PCIe 80GiB (GPU) and $2\times$ AMD EPYC 7T83 64-Core Processor (CPU).

\subsubsection{Baselines} We compare RSP against the following state-of-the-art (SOTA) baselines. \textit{BC} is the simplest imitation learning approach that imitates the offline dataset; \textit{RvS}~\citep{rvs} is an SOTA conditioned imitation learning method that casts offline RL as supervised learning problems; \textit{CQL}~\citep{kumar2020conservative} is an offline RL method that penalizes Q-values for OOD regions; \textit{IQL}~\citep{kostrikov2021iql} is an SOTA in-sample learning offline RL method; \textit{DT}~\citep{chen2021decision} and \textit{TT}~\citep{janner2021offline} are two methods that utilize transformer~\citep{vaswani2017attention} architecture, regarding offline RL problems as sequential modeling problems;\textit{TAP}~\citep{zhang2022efficient} is also a transformer-based method that plans in a latent space encoded by VQ-VAE~\citep{van2017neural}; \textit{Diffuser}~\citep{janner2022planning} builds on top of diffusion model, directly generating the entire fixed-length trajectory and executing the first step.


\definecolor{tblue}{HTML}{1F77B4}
\definecolor{tred}{HTML}{FF6961}
\definecolor{tgreen}{HTML}{429E9D}
\definecolor{thighlight}{HTML}{000000}
\definecolor{cgrey}{rgb}{0.6,0.6,0.6}

\begin{table*}[t!]

\centering
\small
\caption{Comparison of performance in AntMaze navigation, Kitchen manipulation, Mujoco locomotion tasks, and Adroit hand manipulation tasks from the D4RL dataset. AntMaze tasks require long-horizon planning to navigate from the starting point to the goal. Adroit tasks involve high-dimensional action space control of a 24-DoF robot hand. Baselines in {\bf \color{tgreen} teal} utilize expressive models including Transformer or diffusion models. Emphasis is placed on scores within $5\%$ of the maximum for locomotion tasks following, while maximum scores are highlighted for other tasks. \textcolor{cgrey}{$\bm{+G}$} denotes goal-conditioned methods. \textcolor{cgrey}{$\bm{+Q}$} denotes using Q-function as a search heuristic following~\citep{janner2021offline}. We use ``-'' if the original papers do not report scores on the specific tasks. Note that only a few baseline methods report their performance on the challenging Adroit and Kitchen tasks, so we only compare with baselines having reported scores.}
\label{tab:main_results}
\scalebox{0.9}{
\begin{tabular}{ll|ccccc|ccc|c}
\toprule
\multicolumn{1}{c}{\bf \color{black} Dataset} & \multicolumn{1}{c|}{\bf \color{black} Environment} & \multicolumn{1}{c}{\bf \color{black} BC} & \multicolumn{1}{c}{\bf \color{black} RvS \scriptsize{\textcolor{cgrey}{$\bm{(+G)}$}} } & \multicolumn{1}{c}{\bf \color{black} CQL} & \multicolumn{1}{c}{\bf \color{black} IQL} & \multicolumn{1}{c|}{\bf \color{black} HIQL} & \multicolumn{1}{c}{\bf \color{tgreen} DT} & \multicolumn{1}{c}{\bf \color{tgreen} TT \scriptsize{\textcolor{cgrey}{$\bm{(+Q)}$}} } & \multicolumn{1}{c|}{\bf \color{tgreen} TAP \scriptsize{\textcolor{cgrey}{$\bm{(+G)}$}} } & \multicolumn{1}{c}{\bf \color{black} RSP (Ours) \scriptsize{\textcolor{cgrey}{$\bm{(+G)}$}}} \\ 
\midrule
Umaze & AntMaze & $65.0$ & $65.4$ & $74.0$ & $88.2$ & $-$ & $59.2$ & $\textbf{\color{thighlight}100.0}$ & $-$ & $88.2$ \scriptsize{\raisebox{1pt}{$\pm 5.7$}} \\ 
Umaze-Diverse & AntMaze & $55.0$ & $60.9$ & $84.0$ & $66.7$ & $-$ & $53.2$ & $-$ & $-$ & $\textbf{\color{thighlight}92.9}$ \scriptsize{\raisebox{1pt}{$\pm 3.5$}} \\ 
\midrule
Medium-Play & AntMaze & $0.0$ & $58.1$ & $61.2$ & $70.4$ & $84.1$ & $0.0$ & $\textbf{\color{thighlight}93.3}$ & $78.0$ & $91.4$ \scriptsize{\raisebox{1pt}{$\pm 8.8$}} \\ 
Medium-Diverse & AntMaze & $0.0$ & $67.3$ & $53.7$ & $74.6$ & $86.8$ & $0.0$ & $\textbf{\color{thighlight}100.0}$ & $85.0$ & $92.7$ \scriptsize{\raisebox{1pt}{$\pm 3.4$}} \\ 
\midrule
Large-Play & AntMaze & $0.0$ & $32.4$ & $15.8$ & $43.5$ & $86.1$ & $0.0$ & $66.7$ & $74.0$ & $\textbf{\color{thighlight}83.0}$ \scriptsize{\raisebox{1pt}{$\pm 3.7$}} \\ 
Large-Diverse & AntMaze & $0.0$ & $36.9$ & $14.9$ & $45.6$ & $88.2$ & $0.0$ & $60.0$ & $82.0$ & $\textbf{\color{thighlight}86.0}$ \scriptsize{\raisebox{1pt}{$\pm 3.8$}} \\ 
\midrule
Ultra-Play & AntMaze & $0.0$ & $33.2$ & $-$ & $8.3$ & $39.2$ & $0.0$ & $20.0$ & $22.0$ & $\textbf{\color{thighlight}80.3}$ \scriptsize{\raisebox{1pt}{$\pm 4.2$}} \\ 
Ultra-Diverse & AntMaze & $0.0$ & $31.6$ & $-$ & $15.6$ & $52.9$ & $0.0$ & $33.3$ & $26.0$ & $\textbf{\color{thighlight}80.5}$ \scriptsize{\raisebox{1pt}{$\pm 4.1$}} \\ 
\midrule
\multicolumn{2}{c|}{\bf Average} & 15.0 & 48.2 & 50.6 & 51.6 & $-$ & 14.0 & 67.6 & 61.2 & \textbf{\color{thighlight}86.9} \scriptsize{\raisebox{1pt}{$\pm 4.7$}} \\ 
\midrule\midrule
Partial & Kitchen & $38.0$ & $51.4$ & $20.8$ & $46.3$ & $65.0$ & $-$ & $-$ & $-$ & $\textbf{\color{thighlight}76.5}$ \scriptsize{\raisebox{1pt}{$\pm 8.1$}} \\ 
Mixed & Kitchen & $51.5$ & $60.3$ & $24.2$ & $51.0$ & $\textbf{\color{thighlight}67.7}$ & $-$ & $-$ & $-$ & $64.0$ \scriptsize{\raisebox{1pt}{$\pm 20.4$}} \\ 
\midrule
\multicolumn{2}{c|}{\bf Average} & 44.8 & 55.9 & 22.5 & 48.7 & $66.4$ & $-$ & $-$ & $-$ & \textbf{\color{thighlight}70.2} \scriptsize{\raisebox{1pt}{$\pm 14.3$}} \\ 
\bottomrule
\end{tabular}}


\scalebox{0.9}{
\begin{tabular}{ll|cccc|cccc|c}
\toprule
\multicolumn{1}{c}{\bf \color{black} Dataset} & \multicolumn{1}{c|}{\bf \color{black} Environment} & \multicolumn{1}{c}{\bf \color{black} BC} & \multicolumn{1}{c}{\bf \color{black} RvS} & \multicolumn{1}{c}{\bf \color{black} CQL} & \multicolumn{1}{c|}{\bf \color{black} IQL} & \multicolumn{1}{c}{\bf \color{tgreen} Diffuser} & \multicolumn{1}{c}{\bf \color{tgreen} DT} & \multicolumn{1}{c}{\bf \color{tgreen} TT} & \multicolumn{1}{c|}{\bf \color{tgreen} TAP} & \multicolumn{1}{c}{\bf \color{black} RSP (Ours)} \\
\midrule
Medium-Expert & HalfCheetah & $55.2$ & $\textbf{\color{thighlight}92.2}$ & $\textbf{\color{thighlight}91.6}$ & $86.7$ & $88.9$ & $86.8$ & $\textbf{\color{thighlight}95.0}$ & $\textbf{\color{thighlight}91.8}$ & $\textbf{\color{thighlight}92.5}$ \scriptsize{\raisebox{1pt}{$\pm 0.5$}} \\ 
Medium-Expert & Hopper & $52.5$ & $101.7$ & $\textbf{\color{thighlight}105.4}$ & $91.5$ & $103.3$ & $\textbf{\color{thighlight}107.6}$ & $\textbf{\color{thighlight}110.0}$ & $\textbf{\color{thighlight}105.5}$ & $\textbf{\color{thighlight}109.6}$ \scriptsize{\raisebox{1pt}{$\pm 0.7$}} \\ 
Medium-Expert & Walker2d & $\textbf{\color{thighlight}107.5}$ & $\textbf{\color{thighlight}106.0}$ & $\textbf{\color{thighlight}108.8}$ & $\textbf{\color{thighlight}109.6}$ & $\textbf{\color{thighlight}106.9}$ & $\textbf{\color{thighlight}108.1}$ & $101.9$ & $\textbf{\color{thighlight}107.4}$ & $\textbf{\color{thighlight}106.7}$ \scriptsize{\raisebox{1pt}{$\pm 1.0$}} \\ 
\midrule
Medium & HalfCheetah & $42.6$ & $41.6$ & $44.0$ & $\textbf{\color{thighlight}47.4}$ & $42.8$ & $42.6$ & $\textbf{\color{thighlight}46.9}$ & $45.0$ & $42.9$ \scriptsize{\raisebox{1pt}{$\pm 0.2$}} \\ 
Medium & Hopper & $52.9$ & $60.2$ & $58.5$ & $66.3$ & $\textbf{\color{thighlight}74.3}$ & $67.6$ & $61.1$ & $63.4$ & $63.4$ \scriptsize{\raisebox{1pt}{$\pm 2.5$}} \\ 
Medium & Walker2d & $75.3$ & $71.7$ & $72.5$ & $\textbf{\color{thighlight}78.3}$ & $\textbf{\color{thighlight}79.6}$ & $74.0$ & $\textbf{\color{thighlight}79.0}$ & $64.9$ & $\textbf{\color{thighlight}76.6}$ \scriptsize{\raisebox{1pt}{$\pm 1.6$}} \\ 
\midrule
Medium-Replay & HalfCheetah & $36.6$ & $38.0$ & $\textbf{\color{thighlight}45.5}$ & $\textbf{\color{thighlight}44.2}$ & $37.7$ & $36.6$ & $41.9$ & $40.8$ & $40.4$ \scriptsize{\raisebox{1pt}{$\pm 0.4$}} \\ 
Medium-Replay & Hopper & $18.1$ & $73.5$ & $\textbf{\color{thighlight}95.0}$ & $\textbf{\color{thighlight}94.7}$ & $\textbf{\color{thighlight}93.6}$ & $82.7$ & $\textbf{\color{thighlight}91.5}$ & $87.3$ & $88.2$ \scriptsize{\raisebox{1pt}{$\pm 5.6$}} \\ 
Medium-Replay & Walker2d & $26.0$ & $60.6$ & $77.2$ & $73.9$ & $70.6$ & $66.6$ & $\textbf{\color{thighlight}82.6}$ & $66.8$ & $66.9$ \scriptsize{\raisebox{1pt}{$\pm 2.9$}} \\ 
\midrule
\multicolumn{2}{c|}{\bf Average} & 51.9 & 71.7 & \textbf{\color{thighlight}77.6} & \textbf{\color{thighlight}77.0} & \textbf{\color{thighlight}77.5} & 74.7 & \textbf{\color{thighlight}78.9} & 74.8 & \textbf{\color{thighlight}76.4} \scriptsize{\raisebox{1pt}{$\pm 1.7$}} \\ 
\bottomrule
\end{tabular}}


\scalebox{0.9}{
\begin{tabular}{ll|ccc|cc|c}
\toprule
\multicolumn{1}{c}{\bf \color{black} Dataset} & \multicolumn{1}{c|}{\bf \color{black} Environment} & \multicolumn{1}{c}{\bf \color{black} BC} & \multicolumn{1}{c}{\bf \color{black} CQL} & \multicolumn{1}{c|}{\bf \color{black} IQL} & \multicolumn{1}{c}{\bf \color{tgreen} TT} & \multicolumn{1}{c|}{\bf \color{tgreen} TAP} & \multicolumn{1}{c}{\bf \color{black} RSP (Ours)} \\ 
\midrule
Cloned & Pen & $56.9$ & $39.2$ & $37.3$ & $11.4$ & $57.4$ & $\textbf{\color{thighlight}67.9}$ \scriptsize{\raisebox{1pt}{$\pm 6.9$}} \\ 
Cloned & Hammer & $0.8$ & $2.1$ & $2.1$ & $0.5$ & $1.2$ & $\textbf{\color{thighlight}3.6}$ \scriptsize{\raisebox{1pt}{$\pm 3.1$}} \\ 
Cloned & Door & $-0.1$ & $0.4$ & $1.6$ & $-0.1$ & $\textbf{\color{thighlight}11.7}$ & $0.8$ \scriptsize{\raisebox{1pt}{$\pm 0.6$}} \\ 
Cloned & Relocate & $-0.1$ & $-0.1$ & $-0.2$ & $-0.1$ & $-0.2$ & $\textbf{\color{thighlight}0.0}$ \scriptsize{\raisebox{1pt}{$\pm 0.0$}} \\ 
\midrule
Expert & Pen & $85.1$ & $107.0$ & $\textbf{\color{thighlight}133.1}$ & $72.0$ & $127.4$ & $120.9$ \scriptsize{\raisebox{1pt}{$\pm 4.6$}} \\ 
Expert & Hammer & $125.6$ & $86.7$ & $119.5$ & $15.5$ & $127.6$ & $\textbf{\color{thighlight}129.7}$ \scriptsize{\raisebox{1pt}{$\pm 0.9$}} \\ 
Expert & Door & $34.9$ & $101.5$ & $\textbf{\color{thighlight}105.7}$ & $94.1$ & $104.8$ & $105.0$ \scriptsize{\raisebox{1pt}{$\pm 0.3$}} \\ 
Expert & Relocate & $101.3$ & $95.0$ & $106.1$ & $10.3$ & $105.8$ & $\textbf{\color{thighlight}108.1}$ \scriptsize{\raisebox{1pt}{$\pm 0.4$}} \\ 
\midrule
\multicolumn{2}{c|}{\bf Average} & 50.6 & 54.0 & 63.1 & 25.4 & \textbf{\color{thighlight}67.0} & \textbf{\color{thighlight}67.0} \scriptsize{\raisebox{1pt}{$\pm 2.1$}} \\ 
\bottomrule
\end{tabular}}


\end{table*}

\subsubsection{Main results} 
We present evaluation results on D4RL benchmark in Table~\ref{tab:main_results}\footnote{Most baselines didn't report results of the extremely challenging \texttt{Antmaze-Ultra} task so we reproduce BC, DT and RvS and report results of IQL, TT, TAP from TAP paper~\citep{zhang2022efficient}. We collect IQL results on other \texttt{Antmaze-v2} tasks from DOGE~\citep{li2022data} since IQL paper reports scores on \texttt{v0}. We also report self-reproduced results on \texttt{Adroit-Expert} task due to lack of reported scores. Other results are from the original papers.}. The results demonstrate that RSP consistently outperforms or obtains on-par performance compared to other baselines. In particular, RSP achieves unparalleled success in complex \texttt{Antmaze} navigation and \texttt{Kitchen} manipulation tasks that necessitate long-horizon and/or multi-stage planning. For instance, all baselines fail miserably on the extremely challenging \texttt{Antmaze-Ultra} tasks, while RSP can obtain similar success rates as it achieves on the \texttt{Antmaze-Large} tasks. This indicates that the increased difficulty resulting from longer planning horizons has a marginal impact for RSP, demonstrating the superior long-horizon planning capabilities of RSP. Apart from the long-horizon tasks, RSP also demonstrates consistently good performance on Adroit hand manipulation and Mujoco locomotion tasks. On these tasks, Table~\ref{tab:main_results} shows that even with much simpler model architecture, RSP can surprisingly obtain on-par performance results compared to RL as sequential modeling methods that utilize expressive models such as TT, TAP, and Diffuser, as well as TD-based offline RL baselines such as CQL and IQL.

\subsection{Training Cost and Inference Latency}
\label{sec:cost_latency}
To further showcase the advantages of RSP, we monitor the average training time (1M steps) and inference latency (1 step) of different methods on \texttt{Antmaze} tasks and illustrate the connections between training time, inference latency and the performance for different methods in Fig.~\ref{fig:plot_performance}.
All measurements were taken with models trained and inferred on a single GPU in isolation, ensuring no interference from other GPU activities on the server.
We can clearly observe that by leveraging expressive models for sequential modeling, one can sometimes obtain minor improvements in policy performance againts TD-based offline RL approaches. However, these marginal gains come at the expense of exponentially increased training time and inference latency. This inevitably imposes significant computational burdens, restricting the applicability of expressive models in many real-world scenarios. In contrast, RSP with horizon $k=8$ and recursion depth $N=1,4$ obtains the best performance and meanwhile enjoys fast training and minimal inference latency. In particular, RSP ($N=1$) completes the training in approximately 180 seconds and exhibits an inference latency of under 1ms, highlighting its exceptional computational efficiency. 
This sheds light on its potential real-world applicability, given its impressive performance, simplicity, and efficiency.

\subsection{Ablation Studies}\label{sec:ablation}
In this section, we conduct extensive ablation studies on the critical components of RSP including the recursion depth $N$ and the skip-step horizon of the lowest-level sub-goal for policy extraction $k$ (referred to as ``horizon'' hereafter for simplicity). The results are presented in Fig.~\ref{fig:ablation}. 



We first fix the horizon $k=8$ and conduct ablation studies on different recursion depths, i.e. $N=1,2,3,4$. The results in Fig.~\ref{fig:ablation}(a) indicate that a deeper recursion process consistently improves performance. However, performance in some tasks tends to plateau once $N>2$. This is because increasing recursion depth can introduce cumulative errors from each prediction level even though each gets more accurate, which may counterbalance the benefits of RSP for low-level policy learning and result in no further performance improvement. Then we can conclude the following implementation insights: (1) choose minimal recursion depth that achieves saturated performance: $N=2$ is sufficient for most benchmark tasks or those with similar configurations. There is no need to risk RSP performance by unnecessarily increasing depth; (2) advance policy learning as mentioned in Section~\textit{\nameref{sec:discuss}}: Strong policy models may exhibit robustness to slight inaccuracies that accumulate in sub-goal predictions, enabling safe extension of recursion depth and fully harnessing the potential of RSP. Other design considerations in policy formulation, such as how to appropriately condition actions on predicted sub-goals, could further enhance performance.

Next, we fix the recursion depth $N=1$ and ablate on the choices of $k=1,4,16,32$. A larger horizon corresponds to more coarse-grained planning. We observe that the horizon has minimal impact on short-horizon tasks, such as \texttt{Adroit} and \texttt{MuJoCo}, as shown in Fig.~\ref{fig:ablation}(b). However, reducing the horizon significantly leads to performance degradation in long-horizon or multi-stage tasks, such as \texttt{Antmaze} and \texttt{Kitchen}, as seen in Fig.~\ref{fig:ablation}(c) and (d).





\section{Conclusion}
In this study, we propose a novel recursive skip-step planning framework for offline RL, which leverages a set of coarse-grained dynamics models to perform recursive sub-goal prediction, and use a goal-conditioned policy to extract planned actions based on these sub-goals. Our method can smartly bypass the long-horizon compounding error issue in existing sequence modeling-based offline RL methods through hierarchical recursive planning, while still maintaining the advantage of learning in a completely supervised manner. Notably, even using lightweight MLP networks, our method can provide comparable or even better performance as compared to sequence modeling methods that use heavy architectures like Transformers or diffusion models, while providing much better training and inference efficiency. Our work highlights the need to rethink the existing design principles for offline RL algorithms: instead of relying on increasingly heavier models, maybe it's time to introduce more elegant and lightweight modeling schemes to tackle existing challenges in offline RL.

\section{Acknowledgements}
This work is supported by National Key Research and Development Program of China under Grant (2022YFB2502904), National Natural Science Foundation of China under Grant No. 62333015, and Beijing Natural Science Foundation L231014. This work is also supported by Beijing Academy of Artificial Intelligence (BAAI).




\bibliography{aaai25}

\begin{thebibliography}{55}
\providecommand{\natexlab}[1]{#1}

\bibitem[{Ada, Oztop, and Ugur(2023)}]{ada2023diffusion}
Ada, S.~E.; Oztop, E.; and Ugur, E. 2023.
\newblock Diffusion Policies for Out-of-Distribution Generalization in Offline Reinforcement Learning.
\newblock \emph{arXiv preprint arXiv:2307.04726}.

\bibitem[{Ajay et~al.(2022)Ajay, Du, Gupta, Tenenbaum, Jaakkola, and Agrawal}]{ajay2022conditional}
Ajay, A.; Du, Y.; Gupta, A.; Tenenbaum, J.; Jaakkola, T.; and Agrawal, P. 2022.
\newblock Is conditional generative modeling all you need for decision-making?
\newblock \emph{arXiv preprint arXiv:2211.15657}.

\bibitem[{Amos et~al.(2021)Amos, Stanton, Yarats, and Wilson}]{amos2021model}
Amos, B.; Stanton, S.; Yarats, D.; and Wilson, A.~G. 2021.
\newblock On the model-based stochastic value gradient for continuous reinforcement learning.
\newblock In \emph{Learning for Dynamics and Control}, 6--20. PMLR.

\bibitem[{An et~al.(2021)An, Moon, Kim, and Song}]{an2021uncertainty}
An, G.; Moon, S.; Kim, J.-H.; and Song, H.~O. 2021.
\newblock Uncertainty-based offline reinforcement learning with diversified q-ensemble.
\newblock \emph{Advances in Neural Information Processing Systems}.

\bibitem[{Bai et~al.(2021)Bai, Wang, Yang, Deng, Garg, Liu, and Wang}]{bai2021pessimistic}
Bai, C.; Wang, L.; Yang, Z.; Deng, Z.-H.; Garg, A.; Liu, P.; and Wang, Z. 2021.
\newblock Pessimistic Bootstrapping for Uncertainty-Driven Offline Reinforcement Learning.
\newblock In \emph{International Conference on Learning Representations}.

\bibitem[{Chen et~al.(2023)Chen, Lu, Ying, Su, and Zhu}]{chen2022offline}
Chen, H.; Lu, C.; Ying, C.; Su, H.; and Zhu, J. 2023.
\newblock Offline reinforcement learning via high-fidelity generative behavior modeling.
\newblock \emph{International Conference on Learning Representations}.

\bibitem[{Chen et~al.(2021)Chen, Lu, Rajeswaran, Lee, Grover, Laskin, Abbeel, Srinivas, and Mordatch}]{chen2021decision}
Chen, L.; Lu, K.; Rajeswaran, A.; Lee, K.; Grover, A.; Laskin, M.; Abbeel, P.; Srinivas, A.; and Mordatch, I. 2021.
\newblock Decision transformer: Reinforcement learning via sequence modeling.
\newblock \emph{Advances in neural information processing systems}, 34: 15084--15097.

\bibitem[{Cheng et~al.(2024)Cheng, Zhan, Zhang, Lin, Wang, Jiang et~al.}]{cheng2024look}
Cheng, P.; Zhan, X.; Zhang, W.; Lin, Y.; Wang, H.; Jiang, L.; et~al. 2024.
\newblock Look beneath the surface: Exploiting fundamental symmetry for sample-efficient offline rl.
\newblock \emph{Advances in Neural Information Processing Systems}, 36.

\bibitem[{Emmons et~al.(2021)Emmons, Eysenbach, Kostrikov, and Levine}]{emmons2021rvs}
Emmons, S.; Eysenbach, B.; Kostrikov, I.; and Levine, S. 2021.
\newblock RvS: What is Essential for Offline RL via Supervised Learning?
\newblock \emph{arXiv preprint arXiv:2112.10751}.

\bibitem[{Emmons et~al.(2022)Emmons, Eysenbach, Kostrikov, and Levine}]{rvs}
Emmons, S.; Eysenbach, B.; Kostrikov, I.; and Levine, S. 2022.
\newblock RvS: What is Essential for Offline RL via Supervised Learning?
\newblock \emph{International Conference on Learning Representations}.

\bibitem[{Feng et~al.(2022)Feng, Jiang, Yu, Xu, Sun, Wang, Zhan, and Chan}]{feng2022curriculum}
Feng, X.; Jiang, L.; Yu, X.; Xu, H.; Sun, X.; Wang, J.; Zhan, X.; and Chan, W. K.~V. 2022.
\newblock Curriculum Goal-Conditioned Imitation for Offline Reinforcement Learning.
\newblock \emph{IEEE Transactions on Games}.

\bibitem[{Fu et~al.(2020)Fu, Kumar, Nachum, Tucker, and Levine}]{fu2020d4rl}
Fu, J.; Kumar, A.; Nachum, O.; Tucker, G.; and Levine, S. 2020.
\newblock D4rl: Datasets for deep data-driven reinforcement learning.
\newblock \emph{ArXiv preprint}.

\bibitem[{Fujimoto and Gu(2021)}]{fujimoto2021minimalist}
Fujimoto, S.; and Gu, S.~S. 2021.
\newblock A Minimalist Approach to Offline Reinforcement Learning.
\newblock \emph{ArXiv preprint}.

\bibitem[{Fujimoto, Meger, and Precup(2019)}]{fujimoto2019off}
Fujimoto, S.; Meger, D.; and Precup, D. 2019.
\newblock Off-Policy Deep Reinforcement Learning without Exploration.
\newblock In \emph{International Conference on Machine Learning}, 2052--2062.

\bibitem[{Garg et~al.(2023)Garg, Hejna, Geist, and Ermon}]{garg2023extreme}
Garg, D.; Hejna, J.; Geist, M.; and Ermon, S. 2023.
\newblock Extreme Q-Learning: MaxEnt RL without Entropy.
\newblock \emph{arXiv preprint arXiv:2301.02328}.

\bibitem[{Ghosh et~al.(2021)Ghosh, Gupta, Reddy, Fu, Devin, Eysenbach, and Levine}]{ghosh2020learning}
Ghosh, D.; Gupta, A.; Reddy, A.; Fu, J.; Devin, C.~M.; Eysenbach, B.; and Levine, S. 2021.
\newblock Learning to Reach Goals via Iterated Supervised Learning.
\newblock In \emph{International Conference on Learning Representations}.

\bibitem[{Hansen-Estruch et~al.(2023)Hansen-Estruch, Kostrikov, Janner, Kuba, and Levine}]{hansen2023idql}
Hansen-Estruch, P.; Kostrikov, I.; Janner, M.; Kuba, J.~G.; and Levine, S. 2023.
\newblock Idql: Implicit q-learning as an actor-critic method with diffusion policies.
\newblock \emph{arXiv preprint arXiv:2304.10573}.

\bibitem[{Hu et~al.(2023{\natexlab{a}})Hu, Sun, Huang, Guo, Chen, Shen, Sun, Chang, and Tao}]{hu2023instructed}
Hu, J.; Sun, Y.; Huang, S.; Guo, S.; Chen, H.; Shen, L.; Sun, L.; Chang, Y.; and Tao, D. 2023{\natexlab{a}}.
\newblock Instructed Diffuser with Temporal Condition Guidance for Offline Reinforcement Learning.
\newblock \emph{arXiv preprint arXiv:2306.04875}.

\bibitem[{Hu et~al.(2023{\natexlab{b}})Hu, Shen, Zhang, and Tao}]{hu2023graph}
Hu, S.; Shen, L.; Zhang, Y.; and Tao, D. 2023{\natexlab{b}}.
\newblock Graph Decision Transformer.
\newblock \emph{arXiv preprint arXiv:2303.03747}.

\bibitem[{Janner et~al.(2022)Janner, Du, Tenenbaum, and Levine}]{janner2022planning}
Janner, M.; Du, Y.; Tenenbaum, J.; and Levine, S. 2022.
\newblock Planning with Diffusion for Flexible Behavior Synthesis.
\newblock In \emph{International Conference on Machine Learning}, 9902--9915. PMLR.

\bibitem[{Janner et~al.(2019)Janner, Fu, Zhang, and Levine}]{janner2019trust}
Janner, M.; Fu, J.; Zhang, M.; and Levine, S. 2019.
\newblock When to trust your model: Model-based policy optimization.
\newblock \emph{Advances in neural information processing systems}, 32.

\bibitem[{Janner, Li, and Levine(2021)}]{janner2021offline}
Janner, M.; Li, Q.; and Levine, S. 2021.
\newblock Offline Reinforcement Learning as One Big Sequence Modeling Problem.
\newblock \emph{Advances in Neural Information Processing Systems}.

\bibitem[{Jiang et~al.(2022)Jiang, Zhang, Janner, Li, Rockt{\"a}schel, Grefenstette, and Tian}]{jiang2022efficient}
Jiang, Z.; Zhang, T.; Janner, M.; Li, Y.; Rockt{\"a}schel, T.; Grefenstette, E.; and Tian, Y. 2022.
\newblock Efficient planning in a compact latent action space.
\newblock \emph{arXiv preprint arXiv:2208.10291}.

\bibitem[{Konan, Seraj, and Gombolay(2023)}]{konan2023contrastive}
Konan, S.~G.; Seraj, E.; and Gombolay, M. 2023.
\newblock Contrastive decision transformers.
\newblock In \emph{Conference on Robot Learning}, 2159--2169. PMLR.

\bibitem[{Kostrikov et~al.(2021)Kostrikov, Fergus, Tompson, and Nachum}]{kostrikov2021offline}
Kostrikov, I.; Fergus, R.; Tompson, J.; and Nachum, O. 2021.
\newblock Offline Reinforcement Learning with Fisher Divergence Critic Regularization.
\newblock In \emph{International Conference on Machine Learning}, 5774--5783.

\bibitem[{Kostrikov, Nair, and Levine(2021)}]{kostrikov2021iql}
Kostrikov, I.; Nair, A.; and Levine, S. 2021.
\newblock Offline reinforcement learning with implicit q-learning.
\newblock \emph{ArXiv preprint}.

\bibitem[{Kumar et~al.(2019)Kumar, Fu, Soh, Tucker, and Levine}]{kumar2019stabilizing}
Kumar, A.; Fu, J.; Soh, M.; Tucker, G.; and Levine, S. 2019.
\newblock Stabilizing Off-Policy Q-Learning via Bootstrapping Error Reduction.
\newblock In \emph{Advances in Neural Information Processing Systems}, 11761--11771.

\bibitem[{Kumar, Peng, and Levine(2019)}]{kumar2019reward}
Kumar, A.; Peng, X.~B.; and Levine, S. 2019.
\newblock Reward-conditioned policies.
\newblock \emph{arXiv preprint arXiv:1912.13465}.

\bibitem[{Kumar et~al.(2020)Kumar, Zhou, Tucker, and Levine}]{kumar2020conservative}
Kumar, A.; Zhou, A.; Tucker, G.; and Levine, S. 2020.
\newblock Conservative Q-Learning for Offline Reinforcement Learning.
\newblock In \emph{Advances in Neural Information Processing Systems}.

\bibitem[{Levine et~al.(2020)Levine, Kumar, Tucker, and Fu}]{levine2020offline}
Levine, S.; Kumar, A.; Tucker, G.; and Fu, J. 2020.
\newblock Offline reinforcement learning: Tutorial, review, and perspectives on open problems.
\newblock \emph{ArXiv preprint}.

\bibitem[{Li et~al.(2023)Li, Zhan, Xu, Zhu, Liu, and Zhang}]{li2022data}
Li, J.; Zhan, X.; Xu, H.; Zhu, X.; Liu, J.; and Zhang, Y.-Q. 2023.
\newblock When data geometry meets deep function: Generalizing offline reinforcement learning.
\newblock In \emph{International Conference on Learning Representations}.

\bibitem[{Liu et~al.(2021)Liu, Lin, Cao, Hu, Wei, Zhang, Lin, and Guo}]{liu2021swin}
Liu, Z.; Lin, Y.; Cao, Y.; Hu, H.; Wei, Y.; Zhang, Z.; Lin, S.; and Guo, B. 2021.
\newblock Swin transformer: Hierarchical vision transformer using shifted windows.
\newblock In \emph{Proceedings of the IEEE/CVF International Conference on Computer Vision}, 10012--10022.

\bibitem[{Lu et~al.(2023)Lu, Chen, Chen, Su, Li, and Zhu}]{lu2023contrastive}
Lu, C.; Chen, H.; Chen, J.; Su, H.; Li, C.; and Zhu, J. 2023.
\newblock Contrastive Energy Prediction for Exact Energy-Guided Diffusion Sampling in Offline Reinforcement Learning.
\newblock \emph{arXiv preprint arXiv:2304.12824}.

\bibitem[{Lynch et~al.(2020)Lynch, Khansari, Xiao, Kumar, Tompson, Levine, and Sermanet}]{lynch2020learning}
Lynch, C.; Khansari, M.; Xiao, T.; Kumar, V.; Tompson, J.; Levine, S.; and Sermanet, P. 2020.
\newblock Learning latent plans from play.
\newblock In \emph{Conference on robot learning}.

\bibitem[{Lyu et~al.(2022)Lyu, Ma, Li, and Lu}]{lyu2022mildly}
Lyu, J.; Ma, X.; Li, X.; and Lu, Z. 2022.
\newblock Mildly conservative Q-learning for offline reinforcement learning.
\newblock \emph{arXiv preprint arXiv:2206.04745}.

\bibitem[{Mazoure et~al.(2023)Mazoure, Talbott, Bautista, Hjelm, Toshev, and Susskind}]{mazoure2023value}
Mazoure, B.; Talbott, W.; Bautista, M.~A.; Hjelm, D.; Toshev, A.; and Susskind, J. 2023.
\newblock Value function estimation using conditional diffusion models for control.
\newblock \emph{arXiv preprint arXiv:2306.07290}.

\bibitem[{Niu et~al.(2024)Niu, Chen, Liu, Li, Zhou, ZHANG, HU, and Zhan}]{niu2024xted}
Niu, H.; Chen, Q.; Liu, T.; Li, J.; Zhou, G.; ZHANG, Y.; HU, J.; and Zhan, X. 2024.
\newblock x{TED}: Cross-Domain Adaptation via Diffusion-Based Trajectory Editing.
\newblock In \emph{NeurIPS 2024 Workshop on Open-World Agents}.

\bibitem[{Niu et~al.(2022)Niu, Sharma, Qiu, Li, Zhou, Hu, and Zhan}]{niu2022trust}
Niu, H.; Sharma, S.; Qiu, Y.; Li, M.; Zhou, G.; Hu, J.; and Zhan, X. 2022.
\newblock When to trust your simulator: dynamics-aware hybrid offline-and-online reinforcement learning.
\newblock In \emph{Proceedings of the 36th International Conference on Neural Information Processing Systems}, 36599--36612.

\bibitem[{Schmidhuber(2019)}]{schmidhuber2019reinforcement}
Schmidhuber, J. 2019.
\newblock Reinforcement Learning Upside Down: Don't Predict Rewards--Just Map Them to Actions.
\newblock \emph{arXiv preprint arXiv:1912.02875}.

\bibitem[{Van Den~Oord, Vinyals et~al.(2017)}]{van2017neural}
Van Den~Oord, A.; Vinyals, O.; et~al. 2017.
\newblock Neural discrete representation learning.
\newblock \emph{Advances in neural information processing systems}, 30.

\bibitem[{Vaswani et~al.(2017)Vaswani, Shazeer, Parmar, Uszkoreit, Jones, Gomez, Kaiser, and Polosukhin}]{vaswani2017attention}
Vaswani, A.; Shazeer, N.; Parmar, N.; Uszkoreit, J.; Jones, L.; Gomez, A.~N.; Kaiser, {\L}.; and Polosukhin, I. 2017.
\newblock Attention is all you need.
\newblock \emph{Advances in neural information processing systems}, 30.

\bibitem[{Villaflor et~al.(2022)Villaflor, Huang, Pande, Dolan, and Schneider}]{villaflor2022addressing}
Villaflor, A.~R.; Huang, Z.; Pande, S.; Dolan, J.~M.; and Schneider, J. 2022.
\newblock Addressing optimism bias in sequence modeling for reinforcement learning.
\newblock In \emph{international conference on machine learning}, 22270--22283. PMLR.

\bibitem[{Wang et~al.(2023)Wang, Lin, Han, and Lv}]{wang2023offline}
Wang, H.; Lin, Y.; Han, S.; and Lv, K. 2023.
\newblock Offline Reinforcement Learning with Diffusion-Based Behavior Cloning Term.
\newblock In \emph{International Conference on Knowledge Science, Engineering and Management}, 267--278. Springer.

\bibitem[{Wang et~al.(2022)Wang, Zhao, Luo, Ren, Zhang, and Li}]{wang2022bootstrapped}
Wang, K.; Zhao, H.; Luo, X.; Ren, K.; Zhang, W.; and Li, D. 2022.
\newblock Bootstrapped transformer for offline reinforcement learning.
\newblock \emph{Advances in Neural Information Processing Systems}, 35: 34748--34761.

\bibitem[{Wang et~al.(2024)Wang, Xu, Zheng, and Zhan}]{wang2024offline}
Wang, X.; Xu, H.; Zheng, Y.; and Zhan, X. 2024.
\newblock Offline multi-agent reinforcement learning with implicit global-to-local value regularization.
\newblock \emph{Advances in Neural Information Processing Systems}, 36.

\bibitem[{Wang, Hunt, and Zhou(2022)}]{wang2022diffusion}
Wang, Z.; Hunt, J.~J.; and Zhou, M. 2022.
\newblock Diffusion policies as an expressive policy class for offline reinforcement learning.
\newblock \emph{arXiv preprint arXiv:2208.06193}.

\bibitem[{Wu, Tucker, and Nachum(2019)}]{wu2019behavior}
Wu, Y.; Tucker, G.; and Nachum, O. 2019.
\newblock Behavior Regularized Offline Reinforcement Learning.
\newblock \emph{arXiv preprint arXiv:1911.11361.}

\bibitem[{Wu, Wang, and Hamaya(2023)}]{wu2023elastic}
Wu, Y.-H.; Wang, X.; and Hamaya, M. 2023.
\newblock Elastic Decision Transformer.
\newblock \emph{arXiv preprint arXiv:2307.02484}.

\bibitem[{Xiao et~al.(2023)Xiao, Wang, Pan, White, and White}]{xiao2023sample}
Xiao, C.; Wang, H.; Pan, Y.; White, A.; and White, M. 2023.
\newblock The In-Sample Softmax for Offline Reinforcement Learning.
\newblock \emph{arXiv preprint arXiv:2302.14372}.

\bibitem[{Xu et~al.(2023)Xu, Jiang, Li, Yang, Wang, Chan, and Zhan}]{xu2023offline}
Xu, H.; Jiang, L.; Li, J.; Yang, Z.; Wang, Z.; Chan, V. W.~K.; and Zhan, X. 2023.
\newblock Offline RL with No OOD Actions: In-Sample Learning via Implicit Value Regularization.
\newblock \emph{arXiv preprint arXiv:2303.15810}.

\bibitem[{Xu et~al.(2022)Xu, Jiang, Li, and Zhan}]{xu2022policy}
Xu, H.; Jiang, L.; Li, J.; and Zhan, X. 2022.
\newblock A Policy-Guided Imitation Approach for Offline Reinforcement Learning.
\newblock In \emph{Advances in Neural Information Processing Systems}.

\bibitem[{Zhan et~al.(2022)Zhan, Xu, Zhang, Zhu, Yin, and Zheng}]{zhan2022deepthermal}
Zhan, X.; Xu, H.; Zhang, Y.; Zhu, X.; Yin, H.; and Zheng, Y. 2022.
\newblock Deepthermal: Combustion optimization for thermal power generating units using offline reinforcement learning.
\newblock In \emph{Proceedings of the AAAI Conference on Artificial Intelligence}, 4680--4688.

\bibitem[{Zhan, Zhu, and Xu(2022)}]{zhan2022model}
Zhan, X.; Zhu, X.; and Xu, H. 2022.
\newblock Model-based offline planning with trajectory pruning.
\newblock In \emph{International Joint Conference on Artificial Intelligence (IJCAI)}.

\bibitem[{Zhang et~al.(2022)Zhang, Janner, Li, Rockt{\"a}schel, Grefenstette, Tian et~al.}]{zhang2022efficient}
Zhang, T.; Janner, M.; Li, Y.; Rockt{\"a}schel, T.; Grefenstette, E.; Tian, Y.; et~al. 2022.
\newblock Efficient Planning in a Compact Latent Action Space.
\newblock In \emph{International Conference on Learning Representations}.

\bibitem[{Zheng et~al.(2024)Zheng, Li, Yu, Yang, Li, Zhan, and Liu}]{zhengsafe}
Zheng, Y.; Li, J.; Yu, D.; Yang, Y.; Li, S.~E.; Zhan, X.; and Liu, J. 2024.
\newblock Safe Offline Reinforcement Learning with Feasibility-Guided Diffusion Model.
\newblock In \emph{International Conference on Learning Representations}.

\end{thebibliography}

\clearpage
\appendix

\section{Additional Experimental Details}
\label{appendix:exp}
\subsection{Training Implementations}
 We use feed-forward MLPs for all our policies and dynamics models, with two layer neural networks with 1024 units each and ReLU nonlinearities used for all tasks.
 We list the basic hyperparameters in \Cref{tab:hyper}.
 We try to keep the same series of hyperparameters for all the experiment tasks but in \texttt{Kitchen} environment we have to add dropouts and decrease the batch size for less overfitting due to the extremely limit dataset provided by D4RL~\citep{fu2020d4rl}.
 For the recursive dynamics models, [32] denotes ``Recursion step $N=1$" and ``Horizon $k=32$", which implies that we use one dynamics model to predict the probability of the state after 32 steps, i.e. $f(s_{t+32}|s_t,g)$, and extract actions from $\pi(a_t|s_t,s_{t+32},g)$. Similarly, [8,16,32] denotes ``Recursion Depth $N=3$" and ``Horizon $k=8$". 
 For consistency, we adopt [32] across all the tasks in our experiments and achieve superior performance.

 	\begin{table}[ht!]
		\centering
		\caption{Hyperparameters.}
			\begin{tabular}{lc}
				\toprule
				Hyper-parameter & Value\\
				\midrule
                \rowcolor{gray!20}\multicolumn{2}{c}{\textbf{MuJoCo \& Adroit \& Antmaze}} \\
                Dropout & 0.0
                \\
                Batch size & 16384
                \\
                \rowcolor{gray!20}\multicolumn{2}{c}{\textbf{Kitchen}} \\
                Dropout & 0.1
                \\
                Batch size & 256
                \\
                \rowcolor{gray!20}\multicolumn{2}{c}{\textbf{Shared}} \\
                Learning rate & 1e-3
                \\
                Learning rate schedule & cosine
                \\
                Hidden layers & [1024,1024]
                \\
                Recursive skip-step dynamics & [32]\\
				\bottomrule
		\end{tabular}\label{tab:hyper}
	\end{table}

\subsection{Baseline Scores}
Most baselines didn't report results of the extremely challenging \texttt{Antmaze-Ultra} task so we reproduce BC, DT and RvS and report results of IQL, TT, TAP from TAP paper~\citep{zhang2022efficient}. We collect IQL results on other \texttt{Antmaze-v2} tasks from DOGE~\citep{li2022data} since IQL paper reports scores on \texttt{v0}. We also report self-reproduced results on \texttt{Adroit-Expert} task due to lack of reported scores. Other results are from the original papers.

The baseline repositories we use for reproduction are listed below:
\begin{itemize}
    \item BC: https://github.com/Farama-Foundation/D4RL-Evaluations.
    \item RvS: https://github.com/scottemmons/rvs
    \item DT: https://github.com/kzl/decision-transformer
    \item TT: https://github.com/jannerm/trajectory-transformer
    \item TAP: https://github.com/ZhengyaoJiang/latentplan
    \item IQL: https://github.com/ikostrikov/implicit\_q\_learning
\end{itemize}

\end{document}